\documentclass{llncs}
\usepackage[utf8]{inputenc}
\usepackage{amsmath}
\usepackage{graphicx}
\usepackage{amsfonts}
\usepackage{amssymb}
\usepackage[hyphens]{url}
\usepackage[multiple]{footmisc}

\usepackage{mathtools}
\usepackage{multirow}
\usepackage{amssymb}
\usepackage{todonotes}
\usepackage{footmisc}
\usepackage{pifont}
\usepackage{textcomp}
\usepackage{float}
\usepackage{fixltx2e}
\usepackage{caption}
\usepackage{color, colortbl}
\usepackage{xcolor} % might cause issues
\usepackage{booktabs}
\usepackage{subfiles}
\usepackage{listings}
\definecolor{codegreen}{rgb}{0,0.6,0}
\definecolor{codegray}{rgb}{0.5,0.5,0.5}
\definecolor{codepurple}{rgb}{0.58,0,0.82}
\definecolor{codebackcolour}{rgb}{0.95,0.95,0.92}

\lstdefinestyle{mycodestyle}{
  backgroundcolor=\color{codebackcolour},   commentstyle=\color{codegreen},
  keywordstyle=\color{magenta},
  numberstyle=\tiny\color{codegray},
  stringstyle=\color{codepurple},
  basicstyle=\ttfamily\scriptsize,
  breakatwhitespace=false,         
  breaklines=true,                 
  captionpos=b,                    
  keepspaces=true,                 
  numbers=left,                    
  numbersep=5pt,                  
  showspaces=false,                
  showstringspaces=false,
  showtabs=false,                  
  tabsize=2,
  frame=lines
}

\usepackage{svg}

\usepackage[colorlinks=true,linkcolor=blue]{hyperref}   %might cause issues when used before \usepackage{url}

\usepackage{rotating}
\usepackage[T1]{fontenc}
\hypersetup{
    colorlinks,
    linkcolor={red!50!black},
    citecolor={blue!70!black},
    urlcolor={blue!80!black}
}

\definecolor{bittersweet}{rgb}{1.0, 0.44, 0.37}
\definecolor{chamoisee}{rgb}{0.63, 0.47, 0.35}
\definecolor{light-gray}{gray}{0.9}
\definecolor{azure}{rgb}{0.0, 0.5, 1.0}
\definecolor{cadetblue}{rgb}{0.37, 0.62, 0.63}

\usepackage[toc,page]{appendix}
\usepackage{nomencl}
\makenomenclature
\usepackage{etoolbox}
\renewcommand\nomgroup[1]{%
  \item[\Large\bfseries
  \ifstrequal{#1}{N}{Nomenclature}{%
  \ifstrequal{#1}{A}{List of Abbreviations}{}}%
]\vspace{10pt}} % this is to add vertical space between the groups.

\begin{document}

{\let\thefootnote\relax\footnotetext{Copyright \textcopyright\ 2020 for this paper by its authors. Use permitted under Creative Commons License Attribution 4.0 International (CC BY 4.0). CLEF 2020, 22-25 September 2020, Thessaloniki, Greece.}}

\title{A Competitive Deep Neural Network Approach for the ImageCLEFmed Caption 2020 Task}

\author{Marimuthu Kalimuthu \and Fabrizio Nunnari \and Daniel Sonntag}
\institute{
German Research Center for Artificial Intelligence (DFKI) \\ 
Saarland Informatics Campus, 66123 Saarbrücken, Germany\\
%\url{jane@dfki.de}
\{marimuthu.kalimuthu,fabrizio.nunnari,daniel.sonntag\}@dfki.de
}

\maketitle

\begin{abstract}

The aim of ImageCLEFmed Caption task is to develop a system that automatically labels radiology images with relevant medical concepts.  We describe our Deep Neural Network (DNN) based approach for tackling this problem. On the challenge test set of 3,534 radiology images, our system achieves an F1 score of 0.375 and ranks high, 12th among all systems that were successfully submitted to the challenge, whereby we only rely on the provided data sources and do not use any external medical knowledge or ontologies, or pretrained models from other medical image repositories or application domains.\newline


\end{abstract}

\section{Introduction}
\label{sec:intro}
ImageCLEF organises 4 main tasks for the 2020 edition with a global objective of promoting the evaluation of technologies for annotation, indexing, and retrieval of visual data with the aim of providing information access to large collections of images in various usage scenarios and application domains, including medicine~\cite{ImageCLEF20-proceedings}. 

Interpreting and summarizing the insights gained from medical images is a time-consuming task that involves highly trained experts and often represents a bottleneck in clinical diagnosis pipelines. Consequently, there is a considerable need for automatic methods that can approximate this mapping from visual information to condensed textual descriptions. The more image characteristics are known, the more structured are the radiology scans and hence, the more efficient are the radiologists regarding interpretation, see \url{https://www.imageclef.org/2020/medical/caption/} and \cite{ImageCLEFmedConceptOverview2020}.

Recent years have witnessed tremendous advances in deep neural networks in terms of architectures, optimization algorithms, tooling and techniques for training large networks, and handling multiple modalities (e.g., text, images, videos, speech, etc.). In particular, deep convolutional neural networks have proved to be extremely successful image encoders and have thus become the de facto standard for visual recognition~\cite{resnet-he-2016,densenet-huang-2017,vgg16-simonyan-2015}. %Furthermore, there has been several ongoing attempts for integrating different multimedia sources such as \textit{vision} and \textit{language} across the entire spectrum of visual modalities. For a detailed study, we refer the reader to Mogadala et al.~\cite{vl-survey-mogadala-2019}. 
We have been working on machine learning problems in several medical application domains \cite{DBLP:journals/ki/Sonntag20,DBLP:journals/corr/abs-2005-09448,DBLP:journals/artmed/SonntagP19,DBLP:journals/insk/SonntagTZCHRFSG16} in our projects, see \url{https://ai-in-medicine.dfki.de/}. In this paper, we describe how we built a competitive deep neural network approach based on these projects.

The rest of the paper is organized as follows. In Section~\ref{sec:chal}, we formally describe the challenge and its goal. In Section~\ref{sec:meth}, we present some statistics on the dataset and explain the approach we adopted to tackle the challenge. In Section~\ref{sec:expr}, we describe the experiments that we conducted with different architectures and introduce a new loss function that addresses the sparsity problem in ground truth labels. In Section~\ref{sec:res}, results are presented followed by a short discussion. Finally, we summarize our work in Section~\ref{sec:conc} and discuss some future directions. 

\section{The Challenge}
\label{sec:chal}
The overarching goal of ImageCLEFmed Caption challenge is to assist medical experts such as radiologists in interpreting and summarising information contained in medical images. As a first step towards this goal, a simpler task would be to detect as many key concepts as possible, with the goal that these concepts can then be composed into comprehensible sentences, and eventually into medical reports. For full details about the challenge, we refer the reader to Pelka et al.~\cite{ImageCLEFmedConceptOverview2020}.

The challenge has evolved over the years, since its first edition in 2017, to focus only on radiology images in this year's version, and incorporating the lessons learned from previous years. The aim of 2020 ImageCLEFmed Caption challenge is to develop a system that would automatically assign medical concepts to radiology images that were sorted into 7 different categories (see Table~\ref{table:roco-terminology}). More concretely, given an image ($\mathcal{I}$), the objective is to learn a function $\mathcal{F}$ that maps $\mathcal{I}$ to a set of concepts ($\mathcal{C}_{1}$, $\mathcal{C}_{2}$, ..., $\mathcal{C}_{v_\mathcal{I}}$) where $v_\mathcal{I}$ is the number of concepts associated with $\mathcal{I}$. A peculiarity of this challenge is that $v \ll k$, where $k$ is the total number of unique labels.
\begin{align}
    \mathcal{F}: \mathcal{I} \rightarrow (\mathcal{C}_1, \mathcal{C}_2, ..., \mathcal{C}_{v_{\mathcal{I}}})
\end{align}

%\todo[inline]{To revise. We need to specify that the function maps to a subset of size k of the n concepts. But k varies for each image.}

On the challenge data, the ground truth concepts are not known for images in the test set. The only known constraint is that predictions of the test set must be submitted with a maximum of 100 non-repeating concepts per image.

The performance of submissions to the challenge are evaluated on a withheld test set of 3,534 radiology images using the $F1$ score evaluation metric, which is defined as the harmonic mean of \textit{precision} and \textit{recall} values.
\begin{align}
    {F}1 \; &= \; 2 \: * \: \frac{precision \, * \, recall}{precision \, + \, recall}
\end{align}

Firstly, the instance level F1 scores are computed using the \textit{predicted} concepts for images in the test set. Later, an average F1 score is computed over all images in the test set using scikit-learn\footnote{\url{https://scikit-learn.org/stable/modules/generated/sklearn.metrics.f1_score.html}\label{fnote:sklearn-f1-score-url}} library's default \textit{binary} averaging method. This yields the final F1 score for an accepted submission.

All registered teams are allowed for a maximum of 10 submissions. Successful submissions and teams are then ranked based on the achieved F1 scores and results are made publicly available.

\section{Method}
\label{sec:meth}
We provide information about the dataset and some analysis that we performed on it during the exploratory phase. Then, we discuss our learning approach and a suitable data preparation strategy for training our models.

\subsection{Dataset}

All participants are provided with \textit{ImageCLEFmed Caption} dataset which is a subset of the ROCO dataset~\cite{roco-dataset-pelka-2018}. It is a multi-modal, medical images dataset containing radiology images that are each labelled with a set of medical concepts, called as Concept Unique Identifiers (CUIs) in the literature.
As common in many challenges, the provided images are already split into \emph{train}, \emph{validation}, and \emph{test} sets. Images of the first two sets come with ground truth labels, while the \textit{test} set contains only images.

\vspace{-1em}
\begin{table}[h]
\small
    \centering
    \caption{\label{table:roco-terminology} Terminology used in the ROCO dataset.}
    \vspace{0.8em}
    \begin{tabular}{| c | c |}
    \hline  %\toprule
    \rowcolor{cadetblue!55}
        Abbreviation       & Full form \\
    \hline \addlinespace[0.3em]    %\midrule
        DRAN    & DR Angiography      \\
    \rowcolor{gray!15}
        DRCO    & DR Combined modalities in One image     \\
        DRCT    & DR Computerized Tomography  \\
    \rowcolor{gray!15}
        DRMR    & DR Magenetic Resonance    \\
        DRPE    & DR Positron Emission Tomography   \\
    \rowcolor{gray!15}
        DRUS    & DR Ultrasound \\
        DRXR    & DR X-Ray, 2D Tomography \\
    \bottomrule
    \end{tabular}
\end{table}

Moreover, images in each of the splits are sorted into one of the 7 categories as described in tables~\ref{table:roco-terminology} and \ref{table:roco-dataset-stats}; such information can be inferred from the name of the sub-directory containing the images. As we can observe from Table~\ref{table:roco-dataset-stats}, the images are not equally distributed across categories, indicating an imbalance in the dataset. The DRCT category, which contains images captured using Computerized Tomography (CT), has the highest representation, followed by X-ray images (DRXR), while the least number of images (approximately $\frac{1}{40}$th of the images in DRCT) are seen in DRCO category.

\begin{table}[ht]
\small
    \centering
    \caption{\label{table:roco-dataset-stats} Splits and statistics of the ImageCLEFmed Caption dataset.}
    \vspace{1em}
    \begin{tabular}{l | c c c c c c c | c}
    \hline  %\toprule
    %\rowcolor{bittersweet!55}
    \multirow{2}{*}{}        &\multicolumn{7}{c|}{Total Number of Images in the Category of} \multirow{2}{*}{}\\\cline{2-8}
    \multirow{-2}{*}{Split}  &DRAN    &DRCO   &DRCT       &DRMR       &DRPE   &DRUS       &DRXR      &\multirow{-2}{*}{Total}\\\cline{1-9}
    %\hline \addlinespace[0.3em]    %\midrule
        Train               &4,713    &487    &20,031     &11,447     &502    &8,629      &18,944       & 64,753  \\
    \rowcolor{gray!15}
        Val                 &1,132    &73     &4,992      &2,848      &74     &2,134      &4,717        & 15,970  \\
        Test                &325      &49     &1,140      &562        &38     &502        &918          & 3,534  \\
    \hline
        Total               &6,170    &609    &26,163     &14,857     &614    &11,265     &24,579       &84,257  \\
    \bottomrule
    \end{tabular}
\end{table}

Despite having these \textit{category} labels as a meta-information, we could not leverage them during model training due to time constraints. Thus, the relationship between CUIs and image category labels is, for us, still uninvestigated. Possibly, in a follow-up work we will investigate on how to exploit such meta-information to improve classification performance of predictive models, for example by using one of the metadata fusion strategies tested by the authors in~\cite{nunnari20CDMAKE}.

%side by side image
\begin{figure}[t]
  \centering
  \begin{minipage}[b]{0.42\textwidth}
    \includegraphics[width=\textwidth]{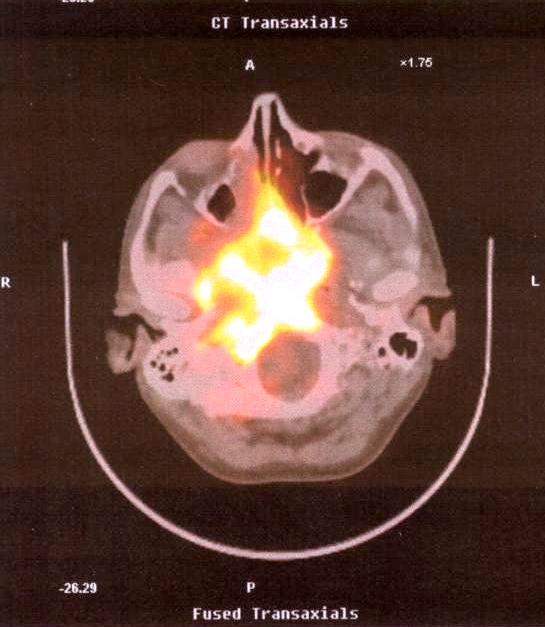}
  \end{minipage}
  \hfill
  \begin{minipage}[b]{0.4\textwidth}
    \includegraphics[width=\textwidth]{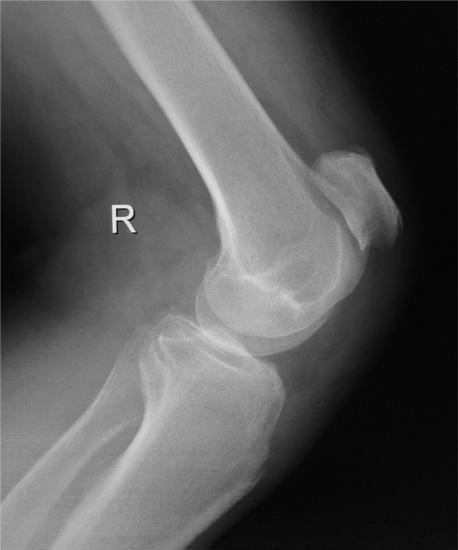}
  \end{minipage}

%side by side table
\vspace{1.2em}\hspace{-2.5em}
\begin{tabular}{ll}
\toprule
\textbf{\textit{CUI}}           &\textbf{\textit{UMLS Term}} (Concept) \\
\toprule %\cmidrule{1-1}
C2951888                        & set of bones of skull  \\
C0037303                        & set of bones of cranium  \\
C0032743                        & tomogr positron emission  \\
C0342952                        & increased basal metabolic rate \\
\bottomrule
\end{tabular}
\hspace{2cm}
\begin{tabular}{ll}
\toprule
\textbf{\textit{CUI}}       &\textbf{\textit{UMLS Term}} \\
\toprule %\cmidrule{1-1}
C0022742                    & knees  \\
C0043299                    & x-ray procedure  \\
C1260920                    & kneel  \\
C0030647                    & bone, patella  \\
\bottomrule
\end{tabular}
\caption{Sample images \& their CUIs from the \textit{train} split.}
\label{fig:img-with-3-cuis}
\end{figure}

\subsection{Data Analysis}
\label{ssec: data-analysis}
Here, we outline some insights that we gained after performing analysis on the CUIs and the category labels meta-information. Furthermore, this section sheds light on the imbalance in the dataset, which is a common problem in many research domains.

Figure~\ref{fig:img-with-3-cuis} provides a conceptual representation of the input, viz. images paired with relevant concepts. In this case, both images are labelled with four CUIs, the descriptions of which are provided by Unified Medical Language System (UMLS)\footnote{\url{https://www.nlm.nih.gov/research/umls/} \label{fnote: umls-data-url}} terms. These terms are depicted in Figure~\ref{fig:img-with-3-cuis} merely for the purpose of understanding since the mapping of CUIs to UMLS terms is not part of the provided dataset.

%CUI freq histogram plots
%ALL-TRAIN-COMBINED
\begin{figure}[t]%[!htbp]
  \centering
    \includegraphics[width=\textwidth]{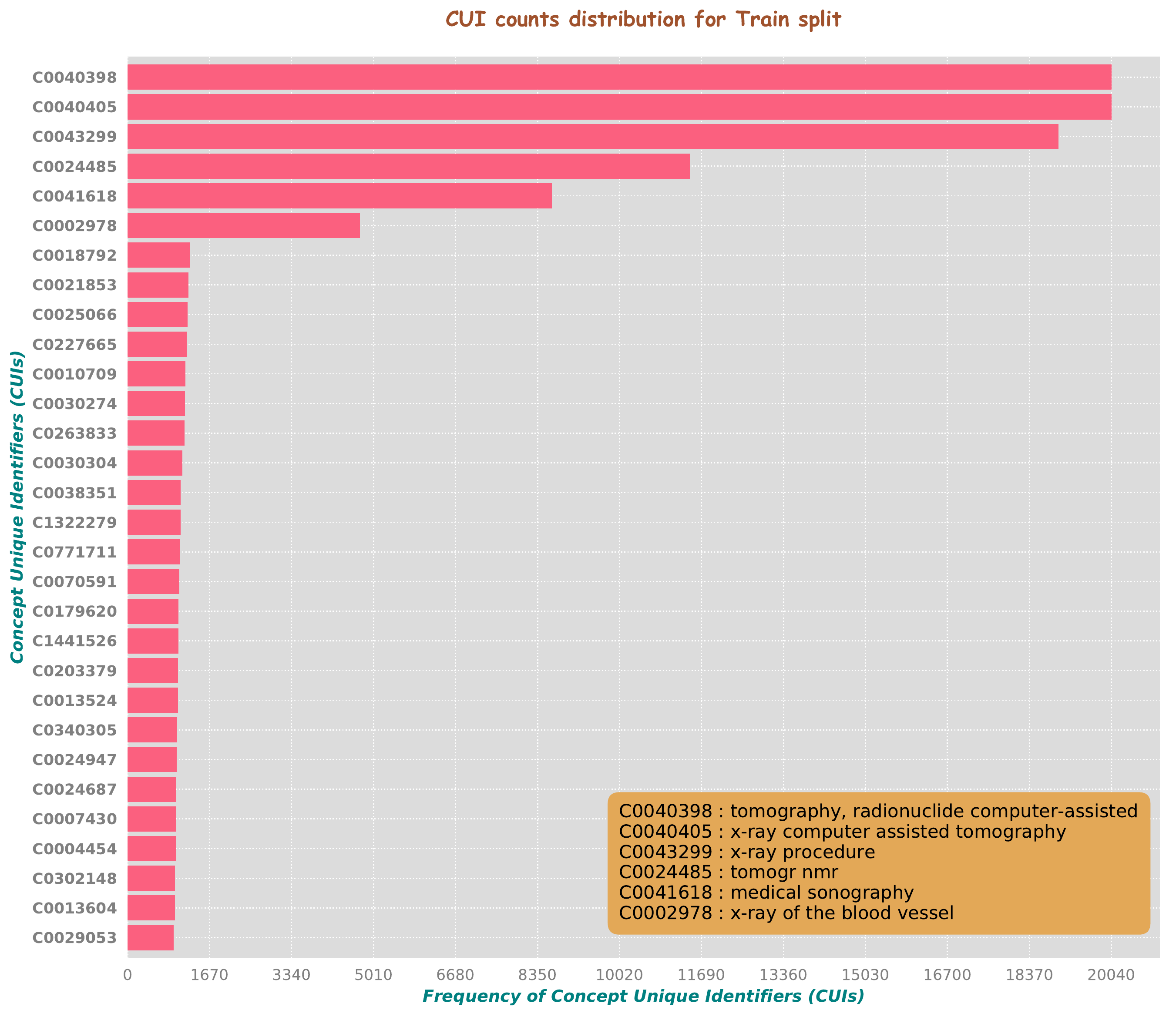}
  \caption{Histogram of top 30 CUIs on the \textit{training} split.}
  \label{fig:cuifreq-all-train}
\end{figure}

%ALL-VAL-COMBINED
\begin{figure}[t]%[!htbp]
  \centering
    \includegraphics[width=\textwidth]{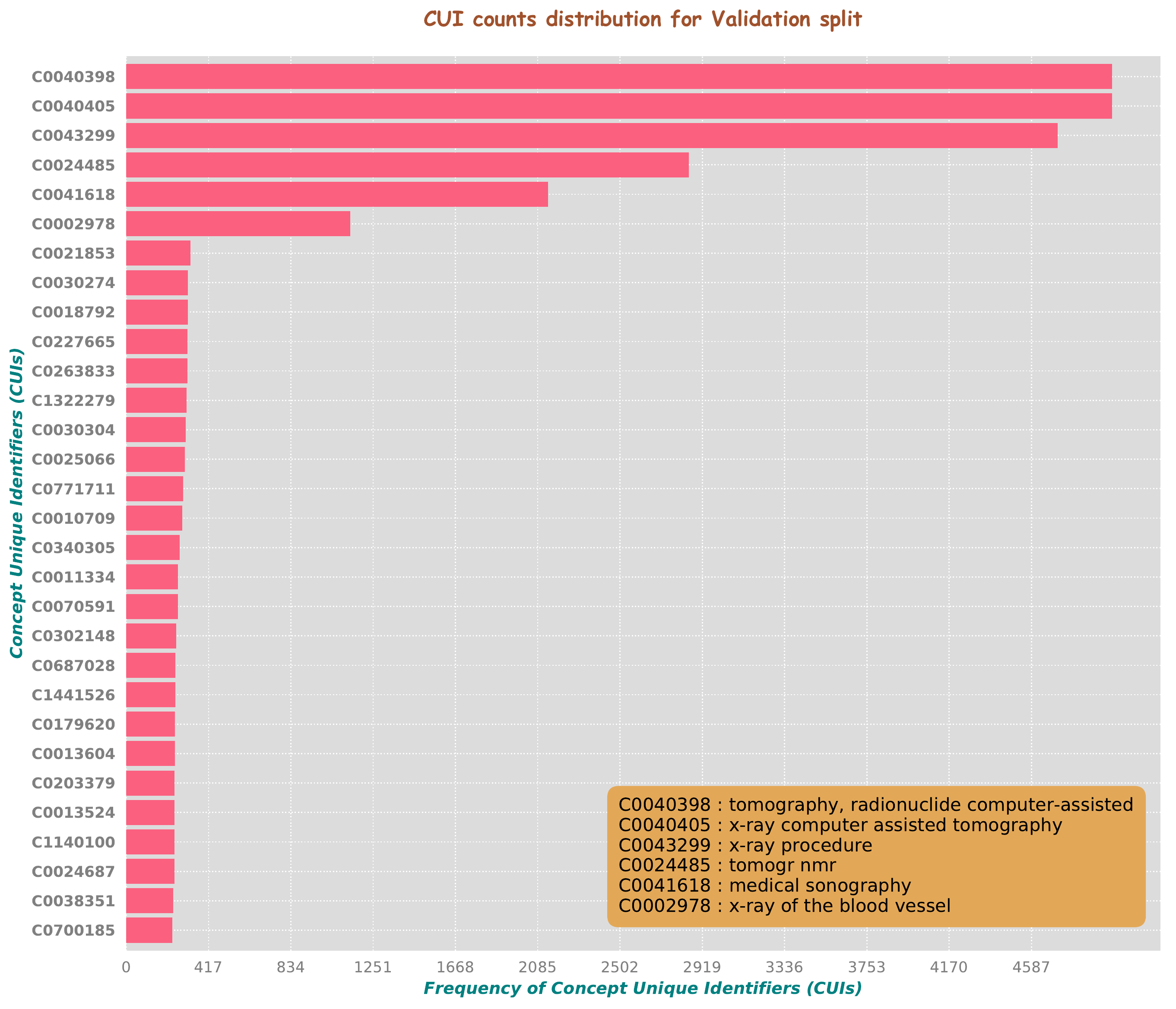}
  \caption{Histogram of top 30 CUIs on the \textit{validation} split.} \label{fig:cuifreq-all-val}
\end{figure}

Figure~\ref{fig:cuifreq-all-train} shows the frequencies of top 30 CUIs on the training split. Two CUIs (\textit{C0040398}, \textit{C0040405}), which both occur around 20k times in the training set, dominate this list and represent the concept ``computer assisted tomography''. Similar behavior is observed on the validation set (see Figure~\ref{fig:cuifreq-all-val}). However, the frequencies of top 2 CUIs in this case are only around 4.6k.

%CUI counts vs. Number of Images histogram plots
%ALL-TRAIN-CATEGORIES-COMBINED
\begin{figure}[t]%[!htbp]
  \centering
    \includegraphics[width=\textwidth]{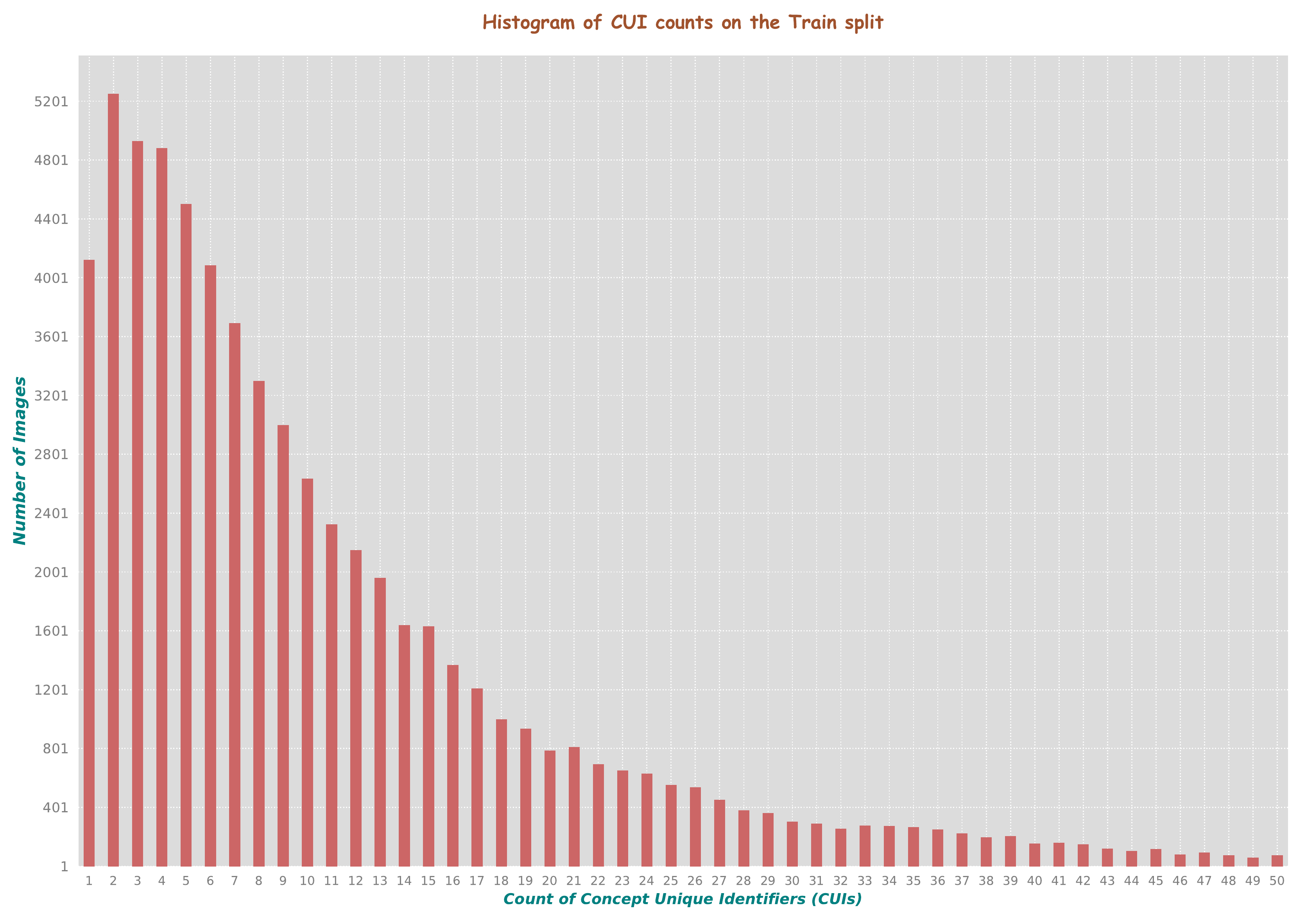}
  \caption{Histogram of CUI counts vs. Images Count on the \textit{training} set after combining all of the 7 categories.}
  \label{fig:cuicounts-vs-images-all-train}
\end{figure}

Figure~\ref{fig:cuicounts-vs-images-all-train} depicts a histogram representation of the number of images and the CUI counts on the training set. For instance, there are around 5,200 images that have exactly two CUIs as ground truth labels. On the contrary, there are only around 100 images that have exactly 50 CUIs as ground truth labels in the provided training set. This histogram is truncated at CUI count 50 (x-axis) for clarity and uncluttered representation.

%ALL-VAL-CATEGORIES-COMBINED
\begin{figure}[t] %[!htbp]
  \centering
    \includegraphics[width=\textwidth]{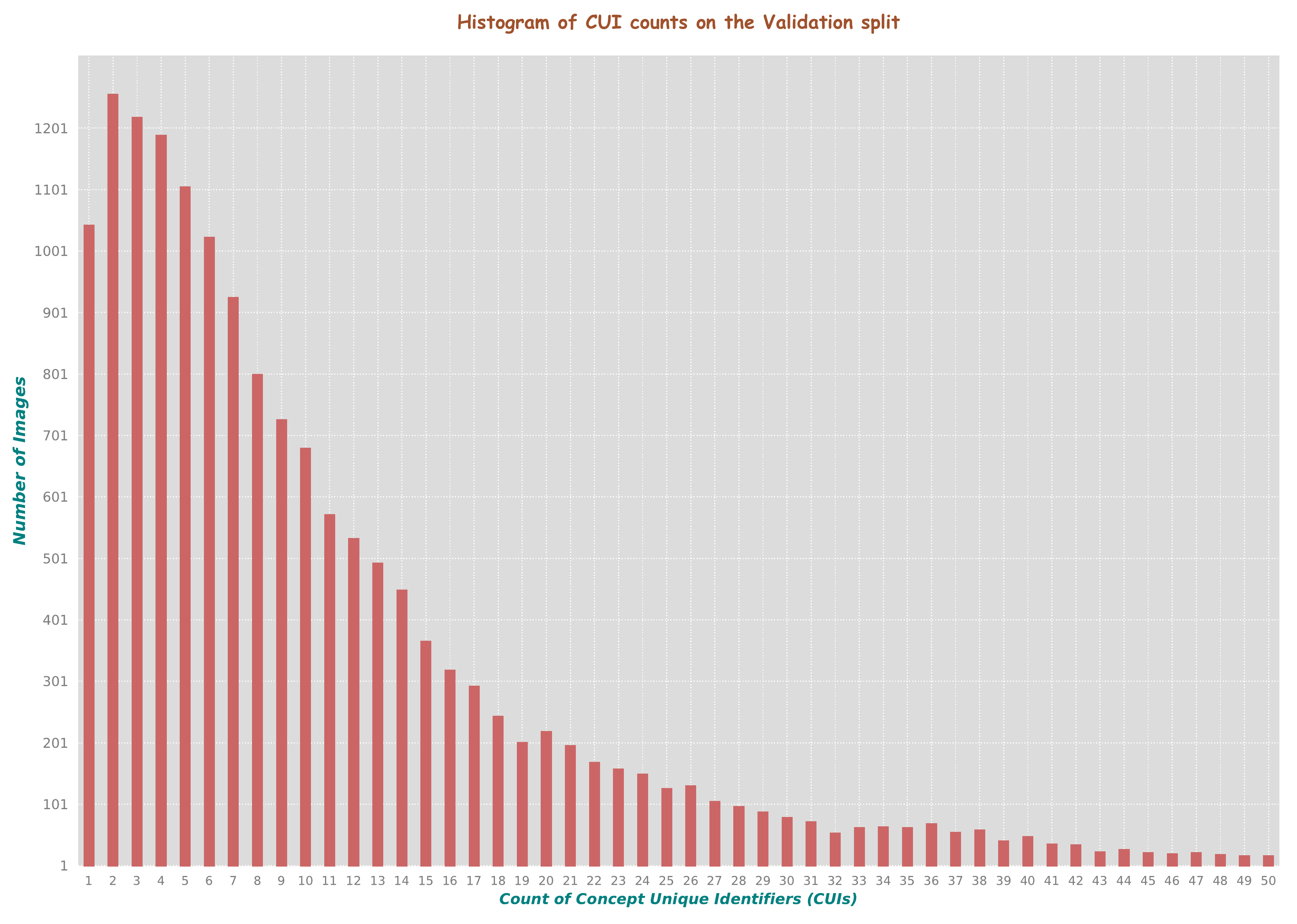}
  \caption{Histogram of CUI counts vs. Images Count on the \textit{validation} set after combining all of the 7 categories.}
  \label{fig:cuicounts-vs-images-all-val}
\end{figure}

In a similar manner, Figure~\ref{fig:cuicounts-vs-images-all-val} shows a histogram representation of the number of images and the CUI counts on the validation set. For instance, there are around 1,300 images that have exactly two CUIs as ground truth labels. On the contrary, there are only around 30 images that have exactly 50 CUIs as ground truth labels in the validation set. This histogram is truncated at CUI count 50 (x-axis) for clarity and uncluttered representation.\\

On the combined training and validation set, there are 80,723 images and 907,718 non-unique CUIs. 
Among them, we counted 3,047 unique CUIs, which were used to build the \textit{label space} in our training objective (see Table~\ref{table:multi-one-hot}).

%\todo[inline]{We should present already a count of the \# of different concepts (3047), and how it was computed.}

Following Tsoumakas et al. \cite{tsoumakas_multi-label_2007}, we compute the \textit{Label Cardinality} (LC) on the combined training and validation set, denoted as $\mathcal{D}$, using the formula:
\begin{align}
    LC(\mathcal{D})  &= \frac{1}{|\mathcal{D}|} \sum_{i=1}^{|\mathcal{D}|} |{Y}_i|
\end{align}

In a similar manner, we compute the \textit{Label Density} (LD) using the following formula:
\begin{align}
    LD(\mathcal{D})  &= \frac{1}{|\mathcal{D}|} \sum_{i=1}^{|\mathcal{D}|} \frac{|{Y}_i|}{|L|}
\end{align}

where $|L|$ is the number of unique labels in our multi-label classification objective. The LC and LD scores on the combined training and validation sets are 11.24 and 0.0037 respectively.

\subsection{Data preparation}

%\todo[inline]{TODO - Mario - Conversion from CLEF format into multi-label vector format}

As a first step in formatting the labels, we convert CUIs associated with images to a format that is suitable as input for neural network learning. Since our objective here is \textit{multi-label} classification, we cannot use simple \textit{one-hot} encoding, as usually done in classification tasks, hence we apply a \textit{multi one-hot} encoding.
An illustration of this representation can be found in Table~\ref{table:multi-one-hot}.
Specifically, we sort the list of CUIs from the unique label set ($k$) in alphabetical order and use the positions of CUIs in the sorted list to mark as $1$ if a specific CUI is associated with the image in question, else as $0$.
After this conversion step, the label set for each image ($\mathcal{I}$) is represented as a single \textit{multi one-hot} vector of fixed size $k$, which is equal to 3,047.

\begin{table}[!ht]
    \small
    \centering
    \caption{\label{table:multi-one-hot} Encoding CUIs using \textit{multi one-hot} representation.}
    \vspace{1em}
    \begin{tabular}{|c | c |}
    \hline %\toprule
    %\rowcolor{cadetblue!45}
        \multirow{2}{*}{}       & Multi One-Hot Encoding \\\cline{2-2}
    %\hline \addlinespace[0.3em]    %\midrule
    %\rowcolor{cadetblue!45}
        \multirow{-2}{*}{Image} & CUI01, CUI02, CUI03, \ldots, CUI3047 \\\cline{1-2}
    %\midrule
    $ROCO2\_CLEF\_76012.jpg$    & ...,0,1,0,0,1,0,0,0,1...      \\
    \rowcolor{gray!15}
    $ROCO2\_CLEF\_05856.jpg$    & ...,0,0,0,0,0,0,0,0,1...     \\
    $ROCO2\_CLEF\_45763.jpg$    & ...,0,1,0,0,1,0,1,0,0...  \\
    \vdots                      & \vdots                    \\
    \hline %\bottomrule
    \end{tabular}
\end{table}

We used only the images from ImageCLEFmed Caption dataset and did not use pre-training on external datasets, or utilize other modalities such as text during model training.

For our experiments, we divided the validation set via random sampling into two equally sized subsets, namely \emph{val1} and \emph{val2}.
We conducted our internal evaluations by always training pairs of models, first using \emph{val1} for validation and \emph{val2} for testing, and then vice-versa. When results were promising, we then submitted our predictions on the test set images by using the model trained with first configuration. Training a third model, based on the validation on the full development set would have been the ideal solution. However, this could not be applied in our case because of time constraints.

\subsection{Learning Approach}

%
% Approach (transfer learning)
The prediction problem for this challenge lays in the category of \emph{multi-label classification} \cite{tsoumakas_multi-label_2007}. It differs from most common classification problems in the fact that each sample of the dataset is simultaneously associated with more than one class from the ground truth label pool.

Technically, when addressing such a problem with deep neural networks, it means that the final classification layer relies on multiple sigmoidal units rather than a single softmax probability distribution.
The last layer of the network contains one sigmoidal unit for each of the target classes, and the association with a true/false result is performed by thresholding the final sigmoid activation value (usually at 0.5).

To address the challenge, we followed a classical transfer learning approach starting from a Convolutional Neural Network (CNN) model pre-trained on an image classification problem, namely ImageNet, because the pre-trained network already offers the ability to detect basic image features, viz. edges, borders, and corners. Then, the final classification stage of the network (i.e., all the layers after the last convolutional layer) is substituted with randomly initialized fully connected layers. Finally, the network is fitted for the new target training set.

In detail, we used VGG16 \cite{vgg16-simonyan-2015}, ResNet50 \cite{resnet-he-2016}, and DenseNet169 \cite{densenet-huang-2017}, all of them pre-trained on ImageNet data used for ILSVRC \cite{krizhevsky_imagenet_2012}.
An example configuration based on VGG16 is shown in listing \ref{lst:VGG16-arch}.
Layers from \texttt{block1\_conv1} to \texttt{block5\_pool} are pre-trained on ImageNet, and unlocked for further training. Remaining layers, from \texttt{flatten} to \texttt{predictions}, are newly instantiated and initialized randomly (where applicable).
The final \texttt{predictions} layer is a dense layer with 3,047 sigmoidal activation units (one per target class).

\begin{lstlisting}[label=lst:VGG16-arch,float=t,caption=An excerpt of the VGG16 architecture used for the multi-label classification task.,captionpos=b,abovecaptionskip=10pt]
_________________________________________________________________
Layer (type)                 Output Shape              Param #   
=================================================================
input_1 (InputLayer)         (None, 227, 227, 3)       0         
_________________________________________________________________
block1_conv1 (Conv2D)        (None, 227, 227, 64)      1792      
_________________________________________________________________
block1_conv2 (Conv2D)        (None, 227, 227, 64)      36928     
_________________________________________________________________
block1_pool (MaxPooling2D)   (None, 113, 113, 64)      0         
_________________________________________________________________
[... 13 more layers ...]
_________________________________________________________________
block5_conv3 (Conv2D)        (None, 14, 14, 512)       2359808   
_________________________________________________________________
block5_pool (MaxPooling2D)   (None, 7, 7, 512)         0         
_________________________________________________________________
flatten (Flatten)            (None, 25088)             0         
_________________________________________________________________
fc1 (Dense)                  (None, 4096)              102764544 
_________________________________________________________________
dropout_1 (Dropout)          (None, 4096)              0         
_________________________________________________________________
fc2 (Dense)                  (None, 4096)              16781312  
_________________________________________________________________
dropout_2 (Dropout)          (None, 4096)              0         
_________________________________________________________________
predictions (Dense)          (None, 3047)              12483559  
=================================================================
Total params: 146,744,103
Trainable params: 146,744,103
Non-trainable params: 0
_________________________________________________________________
\end{lstlisting}

The system was developed in a Python environment using Keras deep learning framework \cite{keras_special_interest_group_keras_nodate} with TensorFlow \cite{martin_abadi_tensorflow_2015} as the backend. For our experiments we used a desktop machine equipped with an 8-core $9^{th}$-gen i7 CPU, 64GB RAM, and NVIDIA RTX TITAN 24GB GPU memory.

\section{Experiments}
\label{sec:expr}
In this section, we describe our experimental procedure, model configuration, and a variety of deep CNN architectures that we tried for achieving the multi-label classification task.

Table \ref{tab:experiments} reports the results of the experiment we conducted throughout the challenge.
Because of time constraints, rather than running a grid search for the best hyper-parameter values, we started from a reference configuration, already successfully used in other past works \cite{nunnari20CDMAKE,nunnari_cnn_2019}.

\begin{table}[t]
    \centering
    \caption{The list of experiments conducted for the challenge.}
    \vspace{1em}
    \scriptsize
    \setlength{\tabcolsep}{1.5pt}
    % Table generated by Excel2LaTeX from sheet 'Sheet2'
\begin{tabular}{r|lllllrll|rlrl|r}
\toprule
\textbf{\#} & \textbf{CNN arch } & \textbf{res} & \textbf{aug.} & \textbf{fc} & \textbf{do} & \multicolumn{1}{l}{\textbf{bs }} & \textbf{loss} & \textbf{lr-red.} & \multicolumn{2}{c}{\textbf{test1}} & \multicolumn{2}{c|}{\textbf{test2}} & \multicolumn{1}{l}{\textbf{score}} \\
      &       & \textbf{(px)} &       & \textbf{layers} &       &       & \textbf{func} &       & \multicolumn{1}{l}{\textbf{ep}} & \textbf{F1} & \multicolumn{1}{l}{\textbf{ep}} & \textbf{F1} & \multicolumn{1}{l}{\textbf{F1}} \\
\midrule
1     & VGG16  & 227   & none  & 2x2k  & 0.5   & 32    & bce   & 0.2/3/loss  & 12    & 0.333 & 22    & 0.335 &  \\
2     & VGG16  & 227   & none  & 2x4k  & 0.5   & 32    & bce   & 0.2/3/loss  & 25    & 0.346 & 26    & 0.336 &  \\
3     & VGG16  & 227   & hflip  & 2x4k  & 0.5   & 32    & bce   & 0.2/5/f1  & 9     & 0.3475 & 9     & 0.3455 & \multicolumn{1}{l}{0.363 } \\
4     & VGG16  & 450   & hflip  & 2x4k  & 0.5   & 24    & bce   & 0.2/5/f1  & \multicolumn{1}{l}{n/a} & crash & 6     & 0.3417 &  \\
\midrule
5     & ResNet50  & 224   & hflip  & 2x4k  & 0.5   & 16    & bce   & nothing  & 3     & 0.3484 & 3     & 0.3487 & \multicolumn{1}{l}{0.365 } \\
6     & DenseNet169  & 224   & none  & 2x4k  & 0.5   & 32    & bce   & nothing  & 8     & 0.3495 & 10    & 0.3450 &  \\
7     & DenseNet169  & 224   & hflip  & 2x4k  & 0.5   & 32    & bce   & nothing  & 5     & 0.3500 & 4     & 0.3463 & \multicolumn{1}{l}{0.360 } \\
\midrule
8     & VGG16  & 227   & hflip  & 3x4k  & 0.5   & 32    & bce   & 0.2/5/f1  & 13    & 0.3465 & 14    & 0.3433 &  \\
\midrule
9    & VGG16  & 227   & hflip  & 2x4k  & 0.5   & 48    & $\overline{F1}*bce$ & 0.2/5/f1  & 11    & 0.3604 & 7     & 0.3606 & \multicolumn{1}{l}{\textbf{0.374 }} \\
10    & VGG16  & 227   & hflip  & 2x4k  & 0.5   & 48    & $\overline{F1}+bce$  & 0.2/5/f1  & 14    & \textbf{0.3636} & 12    & \textbf{0.3632} &  \\
\bottomrule
\end{tabular}%

    \label{tab:experiments}
\end{table}

Together with the base architecture used for convolution (CNN arch) we report: (res) the resolution in pixels of the input images of equal height and width, (aug) the data augmentation strategy, (fc layers) the configuration of final fully-connected stage of the CNN architectures, (do) the dropout value after each fully connected layer, (bs) the batch size used for training, (loss func) the loss function used for optimization, (lr-red) the learning rate reduction strategy (reduction factor/patience/monitored metric). Additionally, we report the best training epoch, based on an early stopping criteria by monitoring the F1 score on the validation set. The last three columns report the F1 scores achieved on the two internal cross-validation sets and finally on the  AIcrowd\footnote{\url{https://www.aicrowd.com/challenges/imageclef-2020-caption-concept-detection}\label{fnote:aicrowd-imageclef-url}} online submission platform.

Other training parameters, common to all configurations are: NAdam optimizer, learning rate = 1e-5, and schedule decay 0.9.
All images were scaled to the input resolution of the CNN using \emph{nearest} filtering, without any cropping.

In the following, we report on the evolution of our tests and obtained results.

%
% VGG16 baseline
\paragraph{\textbf{Experiments 1-4: VGG16 baseline}}

We started with (1) a VGG16 architecture, pretrained on ImageNet, with the last two fully connected (FC) layers configured with n=2048 nodes, each followed by a dropout layer with the dropout probability $p$ set to 0.5.

(2)~We observed an increase in performance by increasing the size of the FC layers to 4096 nodes.

From the first two experiments, it was evident how the loss value (based on binary cross-entropy) could not be effectively used to monitor the validation. The ground truth of each sample, a vector of size 3,047, contains on average about 11 concepts per image (see Section~\ref{ssec: data-analysis}), and only a few images contain more than 50 concepts. Hence, the ground truth matrix is very sparse. As a consequence, the loss function quickly stabilizes into a plateau, as does the accuracy, which saturates to values above 0.9966 after the first epoch. Hence, to better handle early stopping, we implemented a training-time computation of the F1 score.

\begin{figure}[t]
    \centering
    \textbf{Loss (Binary cross-entropy)} \hspace{2.5cm} \textbf{F1 score} \hspace{2cm}~\\
    \vspace{10pt}
    \includegraphics[width=0.43\textwidth]{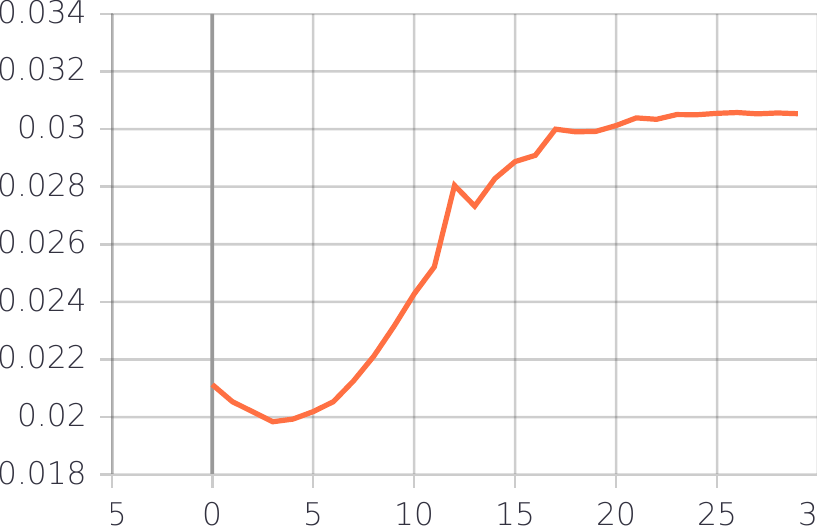}\hspace{1.3cm}
    \includegraphics[width=0.43\textwidth]{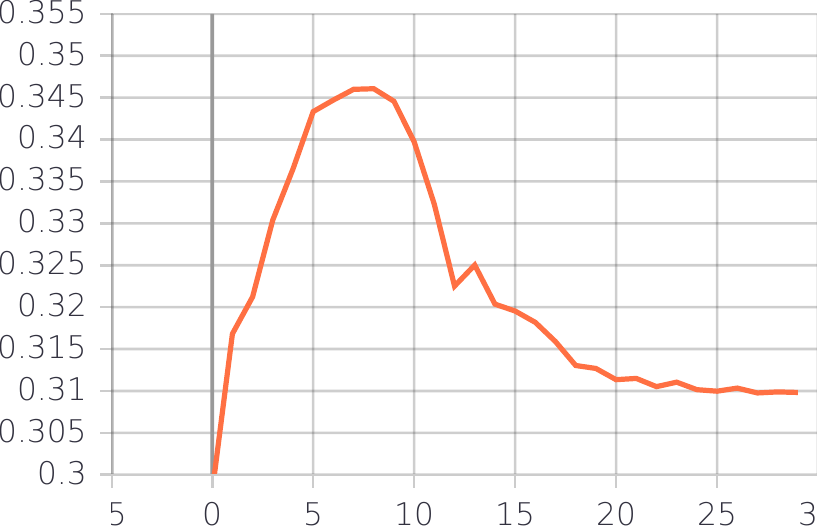}
    \caption{Plots of the metrics computed on the validation set during training: (left) binary cross-entropy and (right) F1 score. Here, epochs are counted from 0. The best validation results are achieved at epoch 9, then the model starts overfitting. }
    \label{fig:training-graphs}
\end{figure}

(3) A further improvement was observed by applying a 2X data augmentation of the input dataset. Each image is provided to the training procedure both as-it-is and flipped horizontally. At the same time, learning rate reduction was applied by monitoring F1 scores on the validation set, rather than the loss values. However, increasing the learning rate happens only after an overfitting occurs. This configuration led to an online evaluation score of 0.363. Figure \ref{fig:training-graphs} shows the evolution of loss values and F1 scores over epochs during model training.

(4) We tried to improve the performance by increasing the size of input images to 450x450 pixels, which forced a reduction of the batch size to 24. We could not observe any significant improvement in the accuracy, suggesting that higher image resolutions do not provide useful details for label selection in our case. 

%
% Better architectures
\paragraph{\textbf{Experiments 5-7: more powerful CNN architectures}}

By using deeper CNN architectures, we could observe a slight improvement in the test accuracy. Indeed, the  ResNet50 architecture (5) led to an F1 score of 0.365 in the online evaluation. The DenseNet169 architecture (6-7) led to higher test values, but the online evaluation was slightly lower (0.360) than the ResNet50 version.

%
% more layers
\paragraph{\textbf{Experiment 8: more layers}}

In order to increase the overall performance, we tried to increase the number of FC layers to 3x4k (8). However, taking as reference the performance of configuration (3), we could not observe a significant improvement by introducing an additional 4k FC layer to the classification stage.

\paragraph{\textbf{Experiments 9-10: a new loss function}}

To further improve performance, we decided to directly optimize for the F1 score evaluation metric. Notice that the F1 score used in the ImageCLEF challenge is computed as an average F1 over the samples (and not over the labels, as more often found in online code repositories\footnote{\url{https://towardsdatascience.com/the-unknown-benefits-of-using-a-soft-f1-loss-in-classification-systems-753902c0105d}}\textsuperscript{,}\footnote{\url{https://www.kaggle.com/rejpalcz/best-loss-function-for-f1-score-metric}}).

We implemented a loss function $\overline{F1}=1-sF1$, where $sF1$ is called the ``soft F1 score''. The $sF1$ is a differentiable version of the F1 function that computes true positives, false positives, and false negatives as continuous sum of likelihood values, without applying any thresholding to round the probabilities to 0 or 1.
The implementation of $\overline{F1}$ is shown in listing \ref{lst:loss-soft-f1}.

\begin{lstlisting}[language=Python,float=t,label=lst:loss-soft-f1,caption=Python implementation of the \textit{soft-F1} based loss function for the Keras environment with TensorFlow backend.,captionpos=b,abovecaptionskip=10pt]
def loss_1_minus_f1(y_true, y_pred):

    import keras.backend as K
    import tensorflow as tf

    # The following is not differentiable.
    # Round the prediction to 0 or 1 (0.5 threshold)
    # y_pred = K.round(y_pred)
    # By commenting, we implement what is called soft-F1.

    # Compute precision and recall.
    tp = K.sum(K.cast(y_true * y_pred, 'float'), axis=-1)
    # tn = K.sum(K.cast((1 - y_true) * (1 - y_pred), 'float'), axis=-1)
    fp = K.sum(K.cast((1 - y_true) * y_pred, 'float'), axis=-1)
    fn = K.sum(K.cast(y_true * (1 - y_pred), 'float'), axis=-1)

    p = tp / (tp + fp + K.epsilon())
    r = tp / (tp + fn + K.epsilon())

    # Compute F1 and return the loss.
    f1 = 2 * p * r / (p + r + K.epsilon())
    f1 = tf.where(tf.is_nan(f1), tf.zeros_like(f1), f1)
    return 1 - K.mean(f1)
\end{lstlisting}

Experiments using the $\overline{F1}$ loss function could not converge. Likely, the problem is due to the fact that the $\overline{F1}$ loss lies in the range $[0,1]$. As such, the gradient search space can be abstractly seen as a huge plateau, just below 1.0, with a solitary hole in the middle that quickly converges to the global minimum.
At the same level of abstraction, we can visualize the binary cross-entropy $bce$ search space as a wide bag, with a large flat surface, just above 0. It is easy to reach the bottom of the bag, i.e., reach a very high binary accuracy due to the sparsity of the labeling, but then we see a very mild slope towards the global minimum at its center.\\

Our intuition is that by combining (multiplying or adding) $\overline{F1}$ and $bce$ results in a search space where $\overline{F1}$ does not affect the identification of inside of the bag, and at the same time helps with the identification of the $\overline{F1}$'s and $bce$'s common global minimum. 
The implementation of the $\overline{F1}*bce$ loss function is straightforward and is presented in listing \ref{lst:loss-f1Xbce}.

\begin{lstlisting}[language=Python,label=lst:loss-f1Xbce,float=t,caption=Python implementation of the loss function combining \textit{soft-F1} score with binary cross-entropy.,captionpos=b,abovecaptionskip=10pt]
def loss_1mf1_by_bce(y_true, y_pred):
    import keras.backend as K

    loss_f1 = loss_1_minus_f1(y_true, y_pred)
    bce = K.binary_crossentropy(target=y_true, output=y_pred, from_logits=False)

    return loss_f1 * bce
\end{lstlisting}

(9) An experiment using VGG16 confirms that the loss function $\overline{F1}*bce$ leads to better results with 0.3604/0.3606 on our tests. This is the configuration that performed best in the online submission (0.374).

Further experiments, e.g., using ResNet50, could not be submitted to the challenge due to time constraints. However, (10) an internal test using VGG16 in combination with the $\overline{F1}+bce$ loss function, led to the best performance in our internal evaluation (0.3636/0.3632).

\section{Results and Discussion}
\label{sec:res}
%
%
%Overall results
The results of our experiments for the ImageCLEFmedical 2020 challenge can be summarized as follows.

The task of concept detection can be modeled as a multi-labeling problem and solved by a transfer learning approach where deep CNNs pretrained on real-world images can be fine-tuned on the target dataset. The multi-labeling is technically addressed by using a sigmoid activation function on the output layer and a label selection by thresholding. A good configuration consists of a VGG16 deep CNN architecture followed by two fully connected layers of 4096 nodes, each followed by a dropout layer with probability $p$ set to 0.5.
Augmenting the training set with horizontally flipped images increases accuracy and also reduces the number of epochs needed for training.
Increasing the resolution of input images does not prove to be useful, while better results are achieved by substituting the convolution stage with a deeper CNN architecture (ResNet50). We noticed that the learning rate reduction has never helped in improving the results.

Using the standard binary cross-entropy loss function leads to competitive results, which significantly increases when it is combined with a soft-F1 score computation. It is worth noticing that when using solely the soft-F1 score as a loss function, the network could not converge and this problem needs further investigation.

% \todo[inline]{Mario, from here on you can comment on the comparison with the other participants.}
% ...

In total, we made five online submissions to the challenge. Table~\ref{tab: results-ranking} presents the F1 scores achieved on the withheld test set and the overall ranking of our team (\textit{iml}) out of 47 successful submissions, as reported by the challenge organizers. In addition, it is worth mentioning that the difference in F1 scores between our best submission and the system that achieved the highest score in the challenge is 0.0195. What percentage of test set images on which our model still needs to achieve correct labels to bridge this gap needs further investigation.

\begin{table}[t]
\small
    \centering
    \caption{\label{tab: results-ranking} F1 scores for submissions by our team \textbf{\textit{`iml'}}.}
    \vspace{1em}
    \begin{tabular}{|l | c | c |}
    \hline  %\toprule
    \rowcolor{brown!45}
        AIcrowd Submission Run     &F1 Score        & Rank\\
    \hline \addlinespace[0.3em]    %\midrule
        imageclefmed2020-test-vgg16-f1-bce-nomissing-iml.txt        & 0.374525478882926      & \textbf{12}       \\
    \rowcolor{gray!15}
        imageclefmed2020-test-vgg16-f1-bce-iml.txt                  & 0.374402134956526      & 13      \\
        imageclefmed2020-test-resnet50-iml.txt                      & 0.365168555515581      & 17      \\
    \rowcolor{gray!15}
        imageclefmed2020-test-vgg16-iml.txt                         & 0.363067945861981      & 18       \\
        imageclefmed2020-test-densenet169-iml.txt                   & 0.360156086299303      & 19       \\
        \addlinespace[0.3em]
    \hline %\bottomrule
    \end{tabular}
\end{table}

\section{Conclusion and Future Directions}
\label{sec:conc}
% future work

In this work we have proposed a deep convolutional neural network based approach for concept detection in radiology images. Our best performance ($12^{th}$ position) is achieved by implementing a new loss function whereby we combined the widely used binary cross-entropy loss together with a differentiable version of the F1 score evaluation metric.

Still several aspects could be investigated to improve the achieved results. For instance, as we can observe from the CUI distribution plots%
%and dataset statistics in Table~\ref{table:roco-dataset-stats}
, there is an imbalance in the dataset. Consequently the model is biased towards predicting the concepts associated with over-represented samples. Our future work will focus on the approaches to combat such type of biases. A straightforward approach would be to undersample the over-represented samples using a query strategy that maximizes informativeness of chosen samples. Such a strategy showed promising results for incremental domain adaptation task in neural machine translation~\cite{incr-da-mario:2019}.

Furthermore, we did not make use of the existing categorization of the images in 7 sub-sets. A straightforward idea would be to train 7 different models, one per category, and rely on their ensemble for a final global classification result. Alternatively, the class identifier might be used as additional metadata information, concatenated to the images' internal features representation in the CNN, and fed to a further shallow neural network for improved classification (see \cite{nunnari20CDMAKE}). Another promising direction to look would be to consider further trends in the integration of vision and language research~\cite{vl-survey-mogadala-2019}.

% references
\bibliographystyle{plain}
\bibliography{references,SkinCare,extra-fab}

\end{document}